%% file: main.tex
\pdfoutput=1
\documentclass[conference]{IEEEtran}
\IEEEoverridecommandlockouts
\usepackage{cite}
\usepackage{amsmath,amssymb,amsfonts}
\usepackage{algorithmicx}
\usepackage{graphicx}
\usepackage{textcomp}
\usepackage{xcolor}
\usepackage{algorithm}
\usepackage{algpseudocode}
\usepackage{bm}
\usepackage{booktabs}
\usepackage{breqn}
\usepackage{multicol}
\usepackage{multirow}
\usepackage{subcaption}
\def\BibTeX{{\rm B\kern-.05em{\sc i\kern-.025em b}\kern-.08em
    T\kern-.1667em\lower.7ex\hbox{E}\kern-.125emX}}

\def\B{\mathcal{B}}
\def\E{\mathbb{E}}
\def\R{\mathbb{R}}
\def\N{\mathcal{N}}
\def\Lb{\mathcal{L}_B}

\begin{document}

\title{Variational Factorization Machines for Preference Elicitation in Large-Scale Recommender Systems}

\author{\IEEEauthorblockN{Jill-Jênn Vie}
\IEEEauthorblockA{\textit{Inria Saclay -- SODA} \\
Palaiseau, France \\
jill-jenn.vie@inria.fr}
\and
\IEEEauthorblockN{Tomas Rigaux}
\IEEEauthorblockA{\textit{Inria Saclay -- SODA} \\
Palaiseau, France \\
tomas.rigaux@inria.fr}
\and
\IEEEauthorblockN{Hisashi Kashima}
\IEEEauthorblockA{\textit{Kyoto University} \\
Kyoto, Japan \\
kashima@i.kyoto-u.ac.jp}
}

\maketitle

\input{content}

\bibliography{biblio.bib}
\bibliographystyle{IEEEtran}

\end{document}

%% file: content.tex
\begin{abstract}
Factorization machines (FMs) are a powerful tool for regression and classification in the context of sparse observations, that has been successfully applied to collaborative filtering, especially when side information over users or items is available. Bayesian formulations of FMs have been proposed to provide confidence intervals over the predictions made by the model, however they usually involve Markov-chain Monte Carlo methods that require many samples to provide accurate predictions, resulting in slow training in the context of large-scale data.
In this paper, we propose a variational formulation of factorization machines that allows us to derive a simple objective that can be easily optimized using standard mini-batch stochastic gradient descent, making it amenable to large-scale data. Our algorithm learns an approximate posterior distribution over the user and item parameters, which leads to confidence intervals over the predictions. We show, using several datasets, that it has comparable or better performance than existing methods in terms of prediction accuracy, and provide some applications in active learning strategies, e.g., preference elicitation techniques.
\end{abstract}

\section{Introduction}

Data collected from a crowd is imperfect: ratings in recommender
systems, outcomes of students over educational exercises, annotations
provided by humans for crowdsourcing tasks. It is crucial to model it
properly in order to provide interesting
recommendations to users, identify misconceptions of students, determine
the true labels. The challenge here relies in modeling both the items
being annotated, and the users annotating them.

In the particular case of recommender systems, we have access to some
ratings provided by users over items, and we want to generalize to new
user-item pairs. Collaborative filtering \cite{Zhou2008} is a famous
technique that consists in learning an embedding for each user and item,
so that the rating can be expressed as a function of the user and item
embeddings. In various applications, side information is also available
on either users or items, such as: the different genres of the movies,
or their actors, directors, etc. The challenge now becomes, how to use
this extra information to improve the recommendations? This is why
factorization machines were developed \cite{Rendle2012}. They generalize
collaborative filtering in the presence of side information.

Models typically encountered in collaborative filtering, such as the
latent factor model, learn point estimates of the user and item
parameters; but it is sometimes more useful to learn distributions over
these parameters, in order to have confidence intervals over predictions
that can guide decision making. In educational applications, this is
particularly useful: what is the uncertainty over the algebra level of
this student? Should I continue to ask them algebra questions, or
should I switch to geometry? On a recommender system, knowing that the predictions are uncertain for a specific user-item pair can be more valuable than just predicting an average grade; using this uncertainty, one can even let
the user control whether they want recommendations with high confidence,
or more risky ones in order to explore more the item database.
This is particularly important when new users come in,
and that we have to identify in few questions their position in latent space to
quickly provide good recommendations to them. This is called preference elicitation \cite{golbandi2011adaptive}.
One way
to do so is to use probabilistic matrix factorization \cite{kim2014scalable}.

However, when we learn these models using Bayesian inference, the
posterior distribution over the parameters is usually intractable, and
we resort to MCMC techniques like Metropolis-Hastings \cite{Cai2010} or
Gibbs sampling \cite{porteous2010bayesian,wang2018confidence}. These
techniques have the advantage to recover the exact distribution
asymptotically, at the cost of many samples. Variational inference
\cite{hoffman2013stochastic,kingma2014stochastic} has been proposed to
compute at a cheaper cost an approximate inference of the posterior.

In this paper, we propose a variational approach to learn factorization
machines. Distributions over the user and item parameters are estimated
easily, and likelihood maximization is done by increasing a lower bound.
By acting directly
on factorization machines, our results naturally extend to the latent
factor models typically encountered in collaborative filtering such as
probabilistic matrix factorization or item response theory.

Our contribution is a simple algorithm for learning Bayesian
factorization machines that works both for classification and regression
tasks. We show using
several datasets of various sparsity that it achieves comparable performance at
predicting unseen pairs than existing methods for regression tasks, and better performance in classification tasks. 

This paper is built as follows. First, we expose related work, then we
present factorization machines, our variational objective, our
algorithm, experiments and results. In particular, we show how those VFMs can be used for querying the preference of users interactively, which is the goal of large-scale recommender systems or intelligent tutoring systems.

\section{Related Work}

Factorization machines (FMs) are a multilinear model of regression or classification using sparse features, described in \cite{Rendle2012}. Under their regression form, FMs can be seen as a generalization of the latent factor model encountered in collaborative filtering. Under their classification form, Vie and Kashima \cite{vie2019knowledge} have shown that FMs can be seen as a generalization of multidimensional item response theory \cite{reckase2009multidimensional}. Variational learning of item response theory models have been proposed at EDM 2020 \cite{wu2020variational} and the methods in the present article can be directly applied to learning multidimensional item response theory models.

Rendle \cite{Rendle2012} describes several
algorithms to train factorization machines implemented in C++ in the libFM package, notably
a Bayesian version where every feature (user, item, or other) has a
Gaussian prior, and Gamma hyperpriors. Using a Markov-chain Monte
Carlo (MCMC) method called Gibbs sampling, they can sample the hyperpriors, the
parameters, then the predictions. One specific advantage of MCMC is that it automatically optimizes the regularization parameter during training, avoiding to perform costly hyper-parameter selection through cross-validation.


Since then, several variants of factorization machines have been proposed, such as
deep factorization machines \cite{guo2017deepfm,Vie2018}, convex factorization
machines \cite{Blondel2015} and higher-order factorization
machines \cite{Blondel2016}. Higher-order FMs consider not only pairwise interactions but $n$-order terms. In all these works, the MCMC version is
recognized as a hard baseline, that sometimes beats deep counterparts
\cite{ferrari2019we,zhu2021open}, which is why it is our principal competitor in this
paper.

To the best of our knowledge, the only independent attempts at
developing variational learning for factorization machines are an
open-source implementation in Chainer \cite{Moody2016} reaching higher RMSE on Movielens 1M than our method; and an unpublished preprint for the regression task only \cite{Saha2018}, with very efficient open-source code in C++, for either full batch or online scenario (batch size of one). As FMs are multilinear in their parameters, \cite{Saha2018} can compute the ELBO objective in closed form, the exact gradients and the best possible updates for each parameter, so their approach is similar to alternating least squares. Terms are cached to optimize the complexity of computation.
However, in the classification task, there is no closed form of the objective, for our choice of distributions and link function.



\section{Factorization Machines}

\def\V{V}

Let \(d \geq 0\) be an integer. Every feature \(k = 1, \ldots, K\), be it a
user, an item, a tag, etc. is parameterized using one scalar bias
\(w_k \in \R\) and an embedding \(\bm{v}_k \in \R^d\). We will note
\(\V\) the matrix having all embeddings \(\bm{v}_1, \ldots, \bm{v}_K\) as columns.

\def\x{\bm{x}}

Each sample \(i = 1, \ldots, n\) is described by a pair \((\bm{x}, y)\)
where \(y\) is a label (say, a rating) and \(\bm{x}\) is a sparse vector
which determines the presence (\(x_k \neq 0\)) or absence (\(x_k = 0\))
of each feature \(k\), see Table~\ref{fm-features}. For example, if
there are \(N_u\) users, \(N_i\) items and no side information, one can
encode the event ``user \(i\) gave item \(j\) 5 stars'' with the pair
\[\left\{\begin{array}{l}
\x = (0, \ldots, 0, \underbrace1_{i}, 0, \ldots, 0, \underbrace1_{N_u + j}, 0, \ldots, 0),\\
y = 5.
\end{array}\right.\]

\begin{table}
\centering
\input{pred-ui}
\caption{Example of sparse features encoded for a factorization machine. Each column of $\bm{x}$ corresponds to either a user or an item.}
\label{fm-features}
\end{table}

\def\w{\bm{w}}
\def\v{\bm{v}}

\begin{figure}
\includegraphics[width=\linewidth]{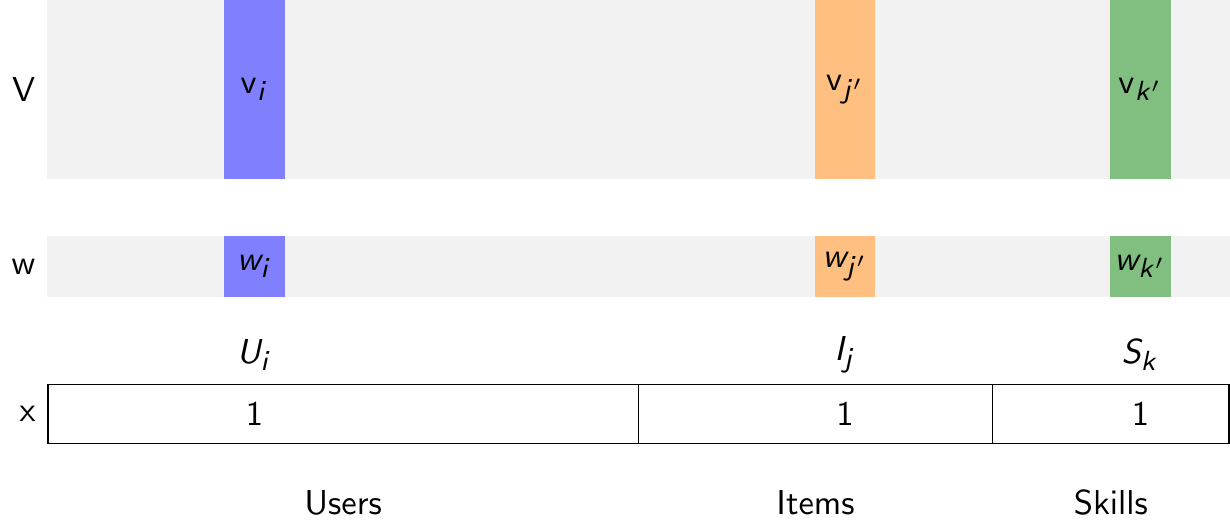}
\includegraphics[width=\linewidth]{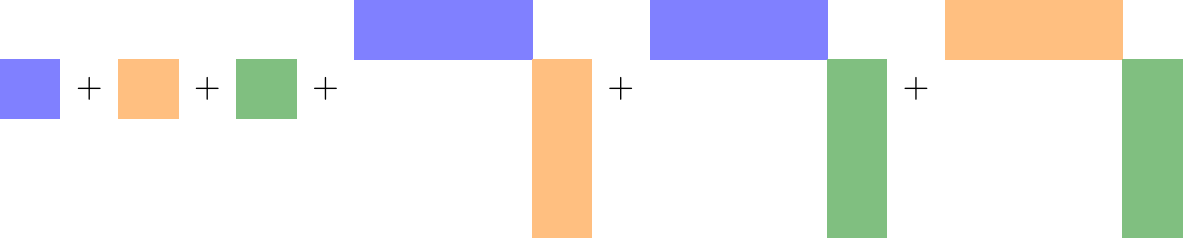}
\caption{Architecture of a FM. Pairwise dot products of embeddings are added to the biases of activated features.}
\label{fm}
\end{figure}

The expression at the core of factorization machines is the following, see also Figure~\ref{fm}:
\begin{align*}
y(\bm{x}) & = w_0 + \sum_{k = 1}^K w_k x_k + \sum_{1 \leq k < \ell \leq K} x_k x_\ell \langle \bm{v}_k, \bm{v}_\ell \rangle\\
& = w_0 + \langle \bm{w}, \bm{x} \rangle + \frac12 \left(|| \V \bm{x} ||^2_2 - \sum_{f = 1}^d || (\V^T)_f \circ \bm{x} ||^2_2 \right)\\
& = w_0 + \langle \bm{w}, \bm{x} \rangle + \frac12 \left(|| \V \bm{x} ||^2_2 - || (\V \circ \V)^T (\bm{x} \circ \bm{x})||^2_2 \right)
\end{align*}

\noindent where \(\circ\) denotes the pairwise product and \(w_0\) is a
global bias. Therefore the parameters of the model are $(w_0, \w, V)$. The last expression is easier to compute as \(\bm{x}\)
is sparse, which also makes it convenient for computing the predictions
of one batch.

In the particular case where there is no side information, we
can recover collaborative filtering: if there are \(N_u\)
users, \(N_i\) items, \(r_{ij}\) is the rating given by user
\(i\) over item \(j\) and \(\bm{x}\) is the concatenation of a
1-hot vector of size \(N_u\) having 1 at position $i$, and a 1-hot vector of size \(N_i\) having 1 at position $j$, we get: 
\[ y(\bm{x}) = w_0 + w_i + w_{N_u + j} + \langle \bm{v}_i, \bm{v}_{N_u + j} \rangle. \]
Therefore, the first $N_u$ entries of $w$ (resp. $V$) contain the user biases (resp. embeddings), the next $N_i$ entries of $w$ (resp. $V$) contain the item biases (resp. embeddings).

\paragraph{Regression}

For a new sample \(\x\) of sparse features, FM will
output a prediction \(\hat{y} \in \N(y(\bm{x}), 1/\alpha)\) where
\(\alpha > 0\) models noise and can be seen as a regularization parameter, and $\N$ is the normal distribution. While optimizing likelihood, we recover a squared loss term.

\paragraph{Classification}

For a new sample \(\x\) of sparse features, FM will
output a prediction \(\hat{y} \in \B(\phi(y(\x))\) where \(\phi\) is a link function and \(\B\) is the Bernoulli distribution. In this paper $\phi = \sigma : x \mapsto 1 / (1 + \exp(-x))$ is the
sigmoid function but in the libFM package, $\phi = \Phi$ is the cumulative density function of the standard normal distribution, because it is more convenient for Gibbs sampling, although very similar in shape. Please note that when $d = 0$, the $V$ term vanishes and we recover logistic
regression when $\phi = \sigma$ and probit regression when $\phi = \Phi$. There is no regularization parameter here, as explained by \cite{doersch2016tutorial} (possibly because the prediction never matches the true label in classification scenarios).

\subsection{Training of FMs}

Training of FMs can be done using standard algorithms such as stochastic
gradient descent (SGD), alternating least squares (ALS) in their regression form, or Gibbs
sampling \cite{Rendle2012} in their Bayesian version where parameters $w_0$,
\(\w\) and \(V\) are random variables. In libFM, the lowest error is usually achieved by training FMs using MCMC.

\section{Variational Factorization Machines (VFM)}

We now detail the variational training of FMs which is our main contribution in this paper.

Let \(\theta\) denote the random variables of a VFM, i.e. $w_0$, \(\bm{w}\) and \(\V\). We will
denote sample \(i\) by its features \(\bm{x}_i\) and outcome \(y_i\).

For regression, the likelihood is
\(p(y_i|\bm{x}_i, \theta) = \N(y(\bm{x}_i), 1/\alpha)\) while for classification, the likelihood is \(p(y_i|\bm{x}_i, \theta) = \B(\sigma(y(\x_i))\) where $\sigma$ is the sigmoid function.

\subsection{Priors}

We use the same prior as libFM:
\[ p(w_k) = \N(\nu^w_{g(k)}, 1/\lambda^w_{g(k)}) \qquad p(v_{kf}) = \N(\nu^{v,f}_{g(k)}, 1/\lambda^{v,f}_{g(k)}) \]
where $g(k) > 0$ is an integer representing the group of feature $k$ (user, item, etc.) and $g(0) = 0$ for the global bias. Rendle uses
hyperpriors \(\nu \sim \N(0, 1), \lambda \sim \Gamma(1, 1) \) where \(\Gamma\) denotes the Gamma
distribution because it is convenient for Gibbs sampling as it is the conjugate of the normal distribution \cite{Rendle2012}. In our case, we just treat $\nu^w, \lambda^w$, $\nu^{v,f}$ and $\lambda^{v,f}$ as parameters.

We found similar results with simpler priors such as $\mathcal{N}(0, 1)$. Saha et al. use $\nu^w = \nu^{v,f} = 0$ \cite{Saha2018}.





\subsection{Approximate posteriors}

\begin{figure}
\includegraphics[width=\linewidth]{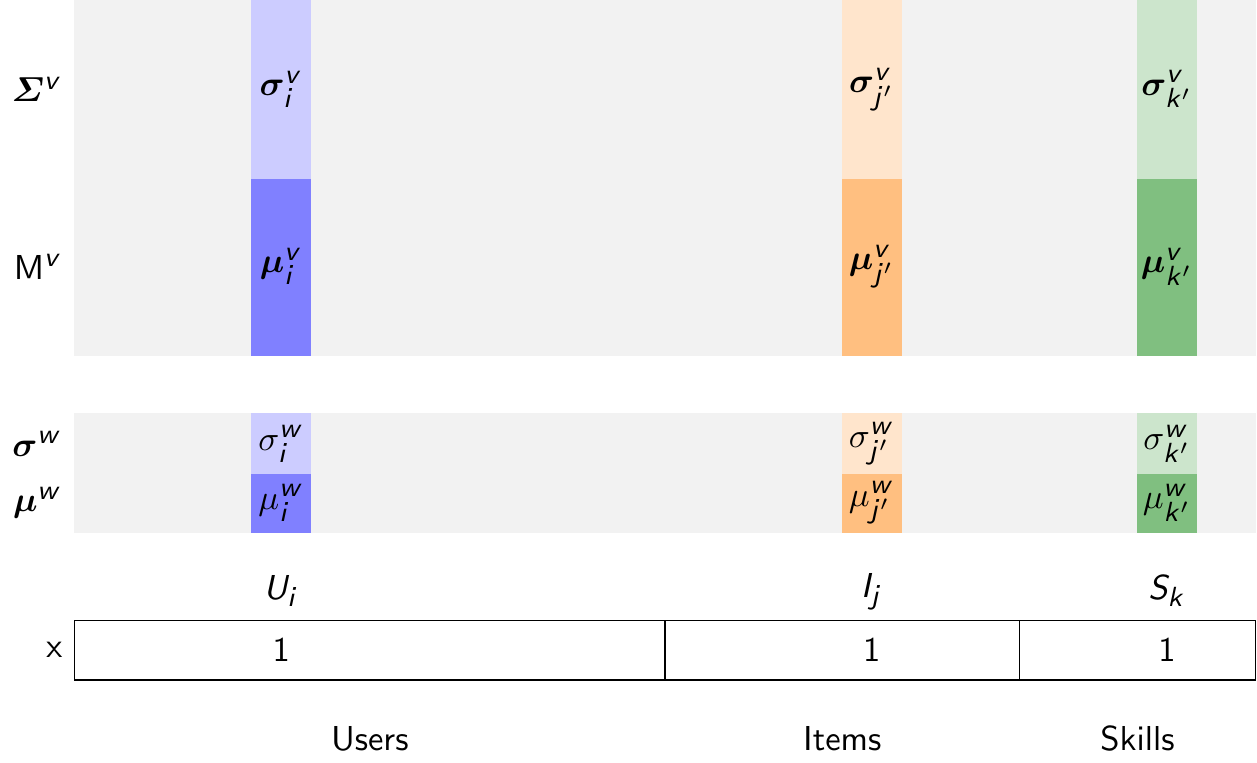}
\caption{Parameters of a VFM, the global bias being omitted for clarity. For each feature we learn a distribution over its bias and a distribution over its embedding. Together, they form an approximate posterior distribution from which we can sample the parameters, and the predictions.}
\label{vfm}
\end{figure}

Our variational approximation:
\[ q(w_k) = \N(\mu^w_k, (\sigma^w_k)^2) \qquad q(v_{kf}) = \N(\mu^{v,f}_k, (\sigma^{v,f}_k)^2) \]

\noindent where \((\mu^w_k, \sigma^w_k, \mu^{v,f}_k,
\sigma^{v,f}_k)\) are parameters that we will learn. Hence, we have
\(2(d + 1)(K + G) + 5\) parameters to estimate representing respectively the posterior of entities, the hyper-parameters related to prior of groups, the global bias, and  and $\alpha$, see Figure~\ref{vfm}.

Sampling a random variable $\theta_k \in \{w_k, \v_k\}$ from the posterior is easy because we just need to do:
\[ \theta_k \leftarrow \mu^\theta_k + \varepsilon \cdot \sigma^\theta_k \]
\noindent for some \(\varepsilon \sim \N(0, 1)\). This is called the reparameterization trick \cite{kingma2014stochastic}.

\subsection{Evidence lower bound objective}

\def\KL{\textnormal{KL}}
\def\y{\bm{y}}

What we would like to increase is the likelihood over all training
samples: \[ \log p(\bm{y}) = \sum_{i = 1}^N \log p(y_i) \]
But it's intractable. So instead we maximize a lower bound
over \(\log p(\y)\): \begin{align*}
\log p(\y) & \geq \sum_{i = 1}^N \underbrace{\E_{q(\theta)} [\log p(y_i|\x_i,\theta)] - \KL(q(\theta)||p(\theta))}_{\textrm{Evidence Lower Bound (ELBO) }}\\
& \quad = \sum_{i = 1}^N \E_{q(\theta)} [ \log p(y_i|\x_i,\theta) ] - \KL(q(w_0)||p(w_0))\\
& \qquad - \sum_{k = 1}^K \KL(q(\theta_k)||p(\theta_k))
\end{align*}
where $q(\theta_k) = q(w_k) q(\v_k) = q(w_k)\prod_{f = 1}^d q(v_{kf}).$

Hopefully, by increasing the Evidence Lower Bound (ELBO), we will
indirectly increase the log-likelihood. This is at the core of
variational inference.

In our case, the ELBO is easy to derive as the KL term is between two Gaussian distributions so it can be computed in closed form. Also all parameters and pairwise independent so everything can be rewritten as a product of univariate Gaussians. 


When we estimate the ELBO for a single batch of samples
\(B \subseteq \{1, \ldots, N\}\), we have to rescale our estimate to alleviate bias. Let $N_k$ (resp. $N_k^B$) be the number of occurrences of feature $k$ in the training set (resp. batch $B$). We
denote \(F(B)\) the set
of unique features that are present in the batch:
\(k \in F(B) \iff N_k^B > 0 \iff \exists i \in B, x_{ik} > 0\).


\begin{align*}
\Lb & = \frac{N}{|B|} \E_{q(\theta)} \Bigg[ \sum_{i \in B} \log p(y_i|\x_i, \theta) \Bigg] \\
& \quad - \frac{K}{\sum_k \alpha_k} \sum_{k \in F(X_B)} \alpha_k \KL(q(\theta_k) || p(\theta_k)) \\
\end{align*}
where $\alpha_k = N_k^B / N_k$ is a rescaling weight for debiasing the ELBO estimate. In practice though, we compute such a weighted combination of KL for each group (users, items) in order to ensure unbiasedness.

The expected log-likelihood can also be written in closed form in the regression case, as the FM model is multilinear in the parameters. This is what \cite{Rendle2012,Saha2018} do to speed up the computations. However, in the classification case, such closed form is not feasible and we have to resort to sampling. $S$ is the number of variational samples, although in most of this paper, $S = 1$.

\def\ll{\frac1{2\alpha} error^2}

\begin{align*}
\Lb & = \frac{N}{|B| S} \sum_{s = 1}^S \sum_{i \in B} \log p(y_i|\x_i, \theta^{(s)})] \\
& \quad - \frac{K}{\sum_k \alpha_k} \sum_{k \in F(B)} \alpha_k \KL(q(\theta_k) || p(\theta_k)) \\
& \theta^{(1)}, \ldots, \theta^{(S)} \sim q(\theta)\\
& \iff \theta_k^{(s)} = \mu^\theta_k + \varepsilon \cdot \sigma^\theta_k \quad \varepsilon \sim \N(0, 1)
\end{align*}




The gradients of \(\Lb\) with respect to parameters
\(\mu_k^w, \sigma_k^w, \mu_k^{v,f}, \sigma_k^{v,f}\) are easy to
compute because the prediction \(y\) is linear in them in the regression case,
log-linear in the classification case. 

We learn the parameters $\mu, \sigma$ and hyper-parameters $\nu, \lambda, \alpha$ using Adam optimizer over
minibatches. We only update the parameters of features that are present
in the batch, which simplifies computation over one batch. Our algorithm
is displayed in Algorithm \ref{algo-vfm}.

\begin{algorithm}
\begin{algorithmic}
\For {each batch $B \subseteq \{1, \ldots, N\}$}
    \State Sample $w_0 \sim q(w_0)$
    \For {$k \in F(B)$ feature involved in batch $B$}
        \State Sample $S$ times $w_k \sim q(w_k)$, $\bm{v}_k \sim q(\bm{v}_k)$
        \State \Comment{Only required if we do not have a closed form}
    \EndFor
    \For {$k \in F(B)$ feature involved in batch $B$}
        \State Update parameters $\mu_k^w, \sigma_k^w, \mu_k^{v,f}, \sigma_k^{v,f}$ to increase ELBO estimate $\Lb$ for this batch
    \EndFor
    \State Update hyper-parameters $\mu_0, \sigma_0, \nu, \lambda, \alpha$
    \State Keep a moving average of the parameters to compute mean predictions
\EndFor
\end{algorithmic}
\caption{Variational Training of FMs}
\label{algo-vfm}
\end{algorithm}

\subsection{Evaluating predictions on a test set}

There are various ways to use the learned distributions to perform prediction. We can either sample from the last iterate of the learned distribution of $w_0$, $\w$ and $V$, or use the last mean (what is called \texttt{last} in experiments), or average iterates of the distribution parameters (what is called \texttt{mean} in the experiments). The latter approach has been denoted as Polyak-Ruppert averaging \cite{polyak1992acceleration} or stochastic weight averaging and acts as a beneficial regularization \cite{neu2018iterate,maddox2019simple,izmailov2018averaging}.




\section{Experiments}

We conducted experiments on real datasets. Given existing entries, the
task consisted in predicting the remaining ones.

\subsection{Tasks and Datasets}

The datasets are described in Table \ref{datasets}. We distinguish three
tasks: regression and classification, which can be respectively understood as matrix completion and binary matrix completion; and preference elicitation.

\paragraph{Regression}

This task is related to collaborative filtering in recommender systems.
Ratings are on an ordinal scale, between 1 and 5 (sometimes including decimal ratings such as 2.5). Predictions on a 20\% test set are
compared using root mean squared error (RMSE):
\[ RMSE = \sqrt{\frac1{|D|} \sum_{i \in D} (y_i - \hat{y}_i)^2}.\]


\paragraph{Classification}

We generated binary Movie100k and Movie1M datasets from the original datasets by setting the outcome to $r_{ij} = 1$ if the rating that user $i$ gave to item $j$ was 4 or 5 stars, 0 otherwise. Predictions on a 20\% test set are compared
using accuracy, area under the ROC curve (AUC) and mean average precision (MAP). This latter metric summarizes a precision-recall curve as a weighted mean of precisions achieved at each threshold, with the increase in recall from the previous threshold used as weight:
$$ MAP = \sum_{n \geq 1} (R_n - R_{n - 1}) P_n \quad \sum_{n \geq 1} R_n - R_{n - 1} = 1 $$
where $P_n$ (resp. $R_n$) is the precision (resp. recall) at the $n$th threshold.

We also made experiments using some side information. The Duolingo dataset for second language acquisition modeling\footnote{https://sharedtask.duolingo.com/2018.html} was used for a competition at the BEA workshop of the NAACL-HLT conference \cite{settles2018second,Vie2018}. It contains the traces of 1,200 English-speaking users learning French by answering exercises among 3 different formats (reverse translate, type what they are hearing, listen and type). The observed outcome of user $i$ learning word $j$ in format of exercise $k$ is 1 if they got it correct, 0 otherwise. Therefore the sparse features are a concatenation of 3 one-hot vectors corresponding to the following groups: user, word, and format of exercise. There are 1.2M million entries but some people try several times the same word over time (only 500k unique user--word pairs), and in this paper we consider the same prediction in this case, as we do not explicitly model learning over time. In order to do it, we can use temporal features as side information, see the literature in knowledge tracing \cite{vie2019knowledge,Vie2018}.



\paragraph{Preference elicitation}

For this final task, we want to guess which items the user knows by selecting few items to ask. We generated the Movie10k dataset for classification from Movielens 25M by
first selecting the 100 movies with the most ratings, then randomly choosing
100 users that had rated at least one but not all of those movies. The
Movie10k dataset is the complete $100 \times 100$ binary matrix $R$ whose
element $r_{ij}$ corresponds to 1 if user $i$ has rated movie $j$, 0 otherwise.



80\% of users are fully observed in the training set for learning item parameters. Once item parameters have been learned, they are frozen and items are separated into interactive training (50\%) and validation sets (50\%). For each of the remaining 20\% of unobserved users in parallel, we select 4 items in the interactive training set for which we reveal whether they have rated those movies or not. Then we update those users' variational parameters, mean and variance, evaluate the predictions on the validation set using ACC, AUC, MAP like in the classification task, then select 4 items again. Another quantity of interest is the reduction in variance for the predictions of items in the validation set.

We compare three strategies for item selection:
\begin{itemize}
    \item random selection in the interactive training set;
    \item mean, which selects the item for which the probability of the FM model is closest to 0.5;
    \item and variance, which selects the item for which the variance of prediction is the highest.
\end{itemize}

\begin{table}
\centering
\input{datasets}
\caption{Datasets used for our experiments.}
\label{datasets}
\end{table}

\begin{figure*}[t]
\centering
\begin{subfigure}{.33\textwidth}
    \centering
    \includegraphics[width=1\linewidth]{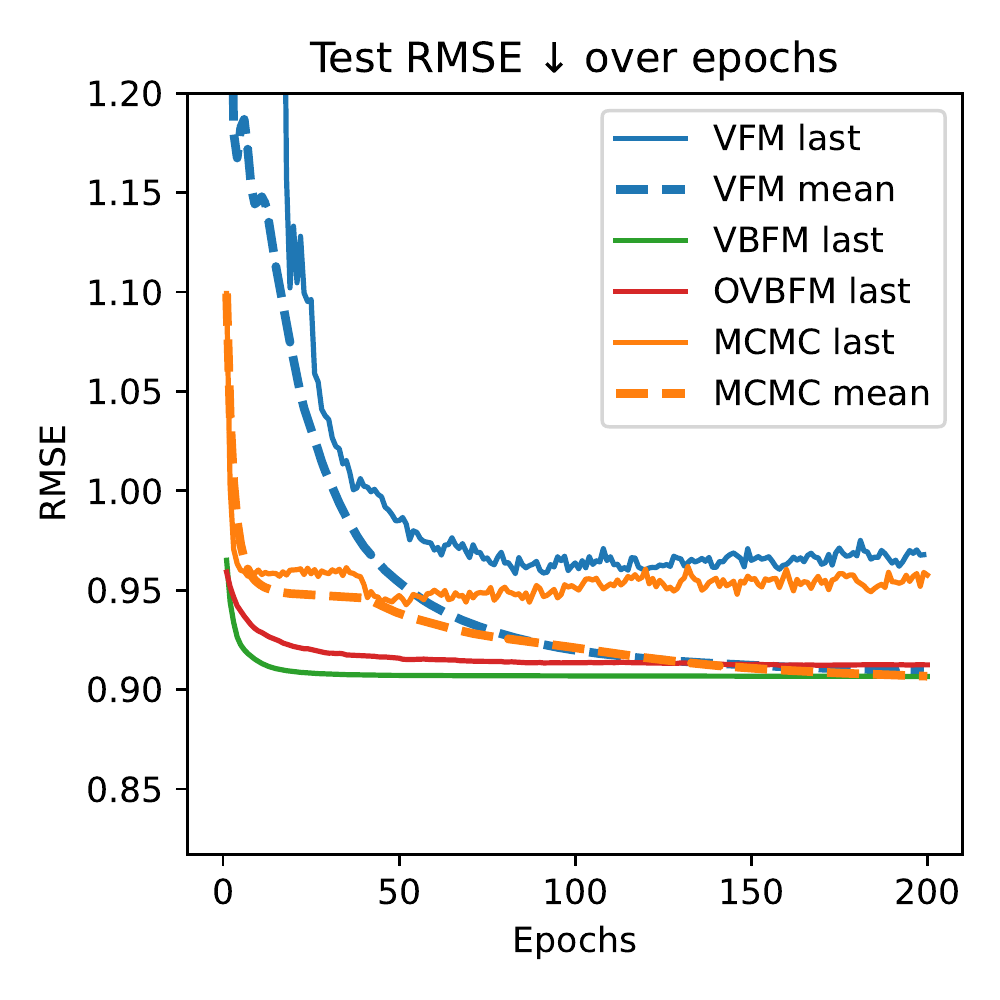}
\end{subfigure}
\begin{subfigure}{.33\textwidth}
    \centering
    \includegraphics[width=1\linewidth]{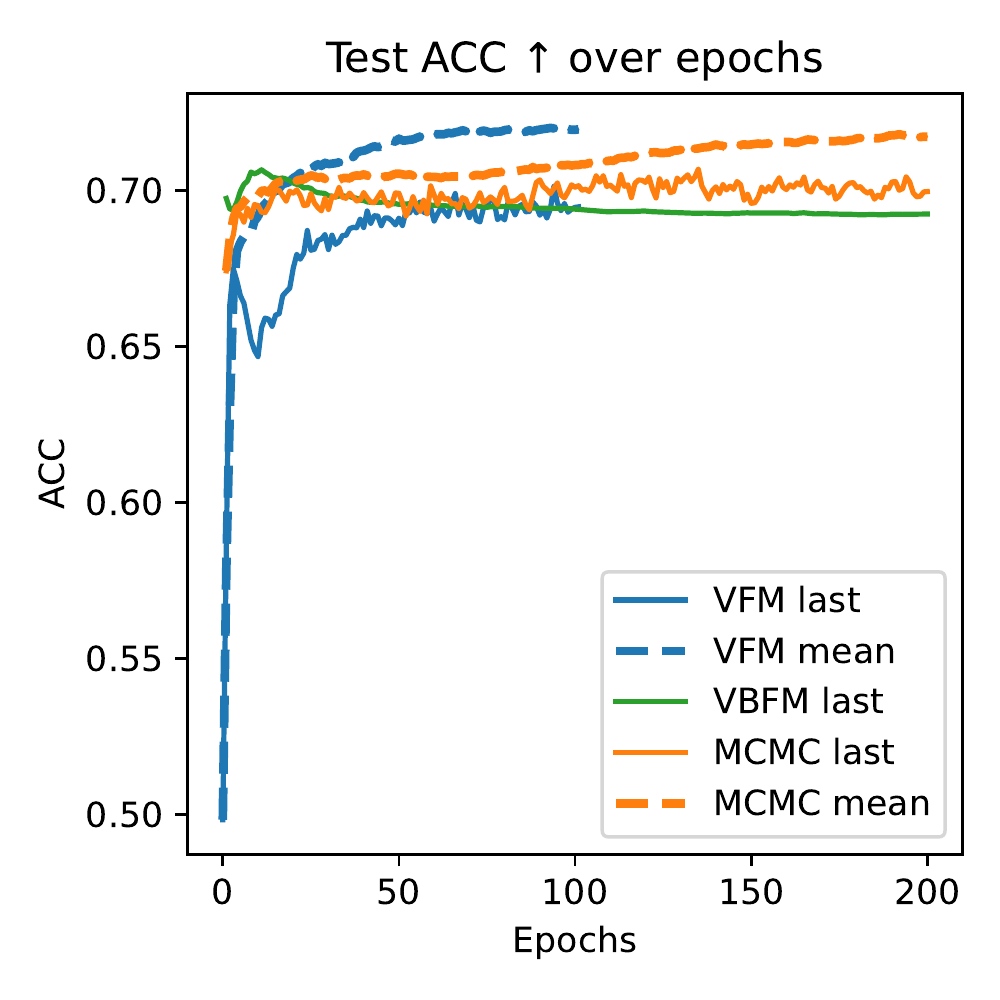}
\end{subfigure}
\caption{Test metrics of all models on Movielens 100k datasets (regression task on the left, classification task on the right). On the right, VFM stopped early at 100 epochs.}
\label{training-movielens}
\end{figure*}

\begin{table*}
\small
\centering
\input{table-sep-4-forced}
\caption{Results on all datasets. First rows are on the regression task, the latter ones on the classification task. The best results are in bold. VBFM and OVBFM is not suited for classification, still it is possible to run VBFM but just to observe accuracy.}
\label{results}
\end{table*}

\subsection{Models}

The main baselines are the libFM\footnote{http://www.libfm.org}
implementation of Gibbs sampling (MCMC) for training FMs
\cite{Rendle2012}, and variational Bayes factorization machines (VBFM) and their online counterpart (OVBFM) from \cite{Saha2018}. OVBFM provides stochastic online training compared to full batch training in VBFM, however the OVBFM implementation from the authors is not expected to work on classification datasets. 


We didn't include any deep model in our benchmarks, as they are often beaten by simpler models in collaborative filtering \cite{ferrari2019we}. Still, we tried DeepFM on Movie100k and got a higher RMSE of 0.93 than the other baselines, although we couldn't perform cross-validation due to time constraints.

\subsection{Framework}

For the regression and classification tasks, 20\% of the training set was used for cross-validating the embedding size $d$ of FMs. We found that among $\{2, 5, 10\}$, $d = 5$ got the best results on the validation set.
In all experiments we
increased our ELBO estimate using the Adam
optimizer \cite{kingma2014adam} over full batch for Movie100k and Movie1M, using learning rate \(\gamma = 0.1\).
We used $S = 1$ variational sample, except for the preference elicitation task where $S = 100$.

$w_0$, $\w$ and $V$ embeddings were initialized using $\N(0, 1)$, precision parameters $\lambda$ initialized with 0.02. To enforce positivity of the standard deviation, we can use absolute value $|\cdot|$ or $\textnormal{softplus}: x \mapsto \log(1 + \exp(x))$. We chose the latter.

For all models considered, all hyperparameters, notably \(\alpha\), were learned during training.
For VFM, the stopping rule for validation was when the validation metric (RMSE or AUC) was worsening for 10 consecutive steps. For training or refit, it was when the ELBO estimate decreased during 4 consecutive steps.
Our implementation, available online\footnote{https://github.com/jilljenn/vae/}, is written in TensorFlow 1.15\footnote{Good old times where TF had more identity than just Keras.} using
TensorFlow Distributions and TensorFlow Probability
\cite{dillon2017tensorflow}. We also provide a simpler implementation in PyTorch using the torch.distributions package.

Using probabilistic programming, we can code the log-likelihood term in the ELBO as $$\texttt{outputs.log_prob(observed).mean()}$$ using the same expression in the classification and the regression case.

\section{Discussion}

\begin{figure*}
\includegraphics[width=\linewidth]{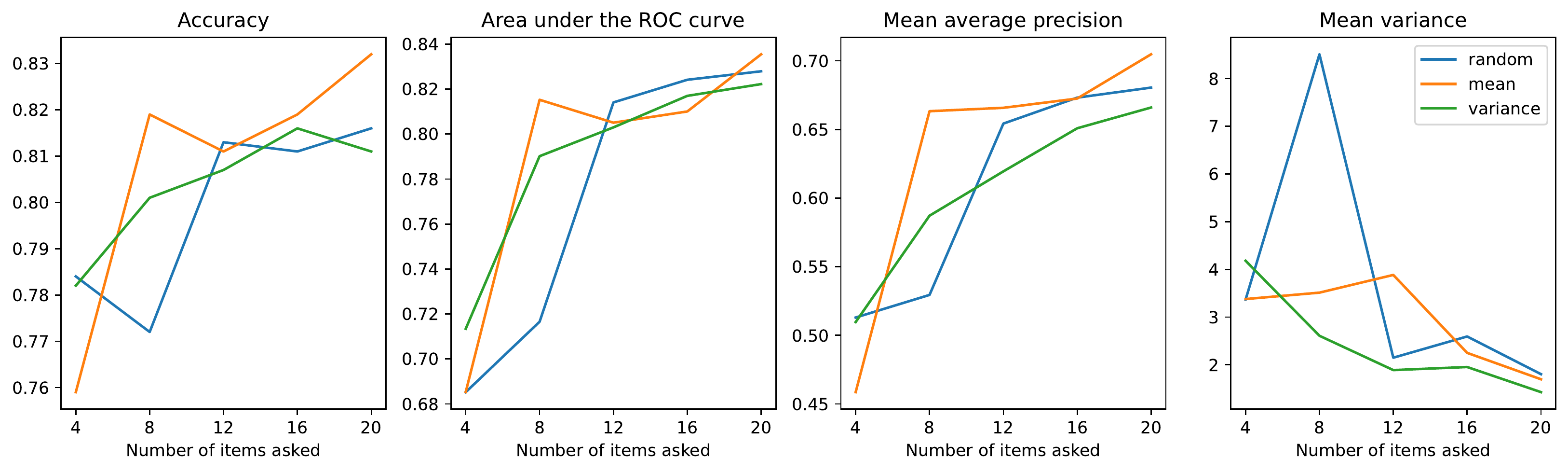}
\caption{Comparing strategies for preference elicitation on the Movie10k dataset.}
\label{pref}
\end{figure*}

\begin{table*}
\normalsize
\begin{tabular}{cccccc} \toprule
\multicolumn{6}{c}{Accuracy}\\
Items & 4 & 8 & 12 & 16 & 20\\ \midrule
Random & \textbf{0.784} & 0.772 & \textbf{0.813} & 0.811 & 0.816\\
Mean & 0.759 & \textbf{0.819} & 0.811 & \textbf{0.819} & \textbf{0.832}\\
Variance & 0.782 & 0.801 & 0.807 & 0.816 & 0.811\\ \bottomrule
\end{tabular} \hfill
\begin{tabular}{cccccc} \toprule
\multicolumn{6}{c}{Area under the ROC curve}\\
Items & 4 & 8 & 12 & 16 & 20\\ \midrule
Random & 0.685 & 0.717 & \textbf{0.814} & \textbf{0.824} & 0.828\\
Mean & 0.685 & \textbf{0.815} & 0.805 & 0.810 & \textbf{0.835}\\
Variance & \textbf{0.713} & 0.790 & 0.803 & 0.817 & 0.822\\ \bottomrule
\end{tabular}
\vspace{5mm}

\begin{tabular}{cccccc} \toprule
\multicolumn{6}{c}{Mean average precision}\\
Items & 4 & 8 & 12 & 16 & 20\\ \midrule
Random & \textbf{0.513} & 0.529 & 0.654 & \textbf{0.673} & 0.680\\
Mean & 0.459 & \textbf{0.663} & \textbf{0.666} & \textbf{0.673} & \textbf{0.705}\\
Variance & 0.510 & 0.587 & 0.619 & 0.651 & 0.666\\ \bottomrule
\end{tabular} \hfill
\begin{tabular}{cccccc} \toprule
\multicolumn{6}{c}{Mean variance of predictions}\\
Items & 4 & 8 & 12 & 16 & 20\\ \midrule
Random & \textbf{3.368} & 8.510 & 2.149 & 2.594 & 1.804\\
Mean & 3.379 & 3.513 & 3.884 & 2.250 & 1.697\\
Variance & 4.181 & \textbf{2.609} & \textbf{1.889} & \textbf{1.954} & \textbf{1.427}\\ \bottomrule
\end{tabular}
\caption{Results on a validation set of different strategies for item selection in an interactive set, on the Movie10k dataset.}
\label{table-interactive}
\end{table*}

The results are reported in Figure \ref{training-movielens} and Table
\ref{results} for the regression and classification tasks, and in Figure \ref{pref} and Table \ref{table-interactive} for the preference elicitation task. For all models, considering the average of weights always leads to better performance than considering only the weights of the last epoch. Such an approach is called stochastic weights averaging \cite{neu2018iterate,maddox2019simple}.

On regression tasks, all models have comparable performance. VBFM is the fastest to converge, thanks to their closed form expressions of gradients. VFM may converge in fewer epochs if we use more variational samples, which induces an artificial reduction in variance but with slower training.

It is natural that VFM may be beaten by MCMC, as the latter does not assume an approximate posterior but samples from the real posterior. Still, on most classification tasks, VFM outperforms MCMC and VBFM in all metrics considered. It also converges faster. Indeed, there is no closed form expression of the gradient updates so VBFM speedup does not hold here.



To optimize the ELBO, we use Adam while VBFM use Robbins-Monro to decay their learning rate \cite{Saha2018}. Indeed, learning rate decays can accelerate convergence in convex scenarios, as we observe it on regression but not on our classification experiments \cite{neu2018iterate,izmailov2018averaging,maddox2019simple,polyak1992acceleration}. 



For the preference elicitation task, we see that most strategies for item selection have comparable performance. The variance strategy can result in higher scores with few items. Selecting items of which the probability of being rated is closest to 0.5 (mean strategy) for the first batch of 4 items does not have a good performance because the resulting items may be collinear and not bring diversity. For this initial batch, random selection or highest variance is more informative. When asking 8 items over 100, the mean strategy performs best, followed by variance. However later, most strategies have comparable performance, with mean still achieving the best metrics after 20 items. 

\section{Future work}

We showed that on large-scale datasets, especially classification datasets, variational training of FMs could
result in faster and better training. Our training relies on batches of data, which
can be used in online inference of FMs, for active learning applications such as preference elicitation. We showed the impact of selecting items of probability close to 0.5 or with high variance in the predictions in order to acquire information efficiently, still many other criteria from information geometry can be used, such as maximizing the trace or determinant of Fisher information. We plan to derive entropy-based strategies, that is, reducing the expected entropy at most at each item. We couldn't do it in this paper as it required to simulate too many possibilities to select the best possible items. However with careful derivation of formulas, such updates can be computed efficiently. 

In this paper, we saw the impact of stochastic weight averaging on performance. Other techniques can be tested, such as putting a stronger regularization on popular features in the dataset, just like ALS-WR (alternating least squares with weighted $\lambda$ regularization) does in collaborative filtering \cite{zhou2008large}.

There sure is a jungle of deep models for factorization but they contain even more hyper-parameters (number of layers, number of neurons) to calibrate. As the input data is sparse, those deep models may overfit. The interested reader may want to look at \cite{shen2017deepctr,zhu2021open,zhu2022bars} for state-of-the-art benchmarks\footnote{https://deepctr-doc.readthedocs.io/en/latest/Features.html}.




Berg et al. have shown that matrix completion can be modeled as message-passing on a graph \cite{berg2018graph}. The variational stochastic gradient
updates in this paper can also be framed this way, as collaborative filtering can be seen as edge prediction in a bipartite graph of users and items. We plan to make more
contributions in this direction, where side information such as movie posters \cite{vie2017using} could be
embedded in the graph and speed up preference elicitation.


We showed that a Gaussian approximation performs well, but it can be
replaced with other, more complex families using normalizing flows or even stable diffusion \cite{rombach2022high}, as
long as the Jacobian is easy to compute \cite{Dinh2016}.

\section{Conclusion}

In this paper we showed how factorization machines could be trained
easily by optimizing a simple variational objective. By recovering an
approximation of the posterior distribution over the parameters of the
model, one can identify which features need extra data points. Another advantage is that hyperparameter optimization is automatically done during training. This work
opens the door to new techniques of active learning, or upper confidence bound algorithms from multi-armed
bandits, because now reward functions can be expressed in function of
the uncertainty of the model. We hope to see many applications in adaptive education and interactive recommender systems.

\section{Acknowledgements}

We thank Akash Srivastava for friendly and enlightened discussions in Tokyo. This work was started while Jill-Jênn Vie was a researcher in Kyoto University funded by RIKEN AIP; and completed while the first two authors were visiting the third author in Kyoto U, funded by the OPALE associated team between Inria and Kyoto U. This work was supported by JST CREST Grant Number JPMJCR21D1.

%% file: pred-ui.tex
\begin{tabular}{lrrrrrrrr}
\toprule
& \multicolumn{6}{c}{Sparse features $\bm{x}$} & \multirow{2}{*}{$\bm{y}$}\\
\cmidrule{2-4} \cmidrule{5-7}
{} &  $U_0$ &  $U_1$ &  $U_2$ &  $I_0$ &  $I_1$ &  $I_2$ &      \\
\midrule
User 1 Item 1: 5/5  &  0 &  1 &  0 &  0 &  1 &  0 &      5 \\
User 1 Item 2: 4/5 &  0 &  1 &  0 &  0 &  0 &  1 &       4 \\
User 2 Item 1: 2/5 &  0 &  0 &  1 &  0 &  1 &  0 &       2 \\
User 2 Item 1: 2/5  &  0 &  0 &  1 &  0 &  1 &  0 &    2 \\
User 2 Item 2: 5/5 &  0 &  0 &  1 &  0 &  0 &  1 &     5 \\
\bottomrule
\vspace{-5pt}
\end{tabular}

%% file: datasets.tex
\begin{tabular}{llrrrr}
\toprule
Task & Dataset &  \#users &  \#items &  \#entries & Sparsity \\
\midrule
Regression & Movie100k &     944 &    1683 &    100000 &    0.937 \\
           & Movie1M   &    6041 &    3707 &   1000209 &    0.955 \\
\midrule
Classification & Movie10k   &     100 &     100 &      10000 &    0 \\
 & Movie100k &     944 &    1683 &    100000 &    0.937 \\
           & Movie1M   &    6041 &    3707 &   1000209 &    0.955 \\
    & Duolingo & 1213 & 2416 & 1199732 & 0.828\\
\bottomrule
\vspace{-5pt}
\end{tabular}

%% file: table-sep-4-forced.tex
\begin{tabular}{cccccc}
\toprule
        &                &                 FM MCMC \cite{Rendle2012} &                     VFM &     VBFM \cite{Saha2018} & OVBFM  \\
\midrule
Movie100k & RMSE &  \textbf{0.906} &  \textbf{0.906} &  0.907 & 0.912  \\
Movie1M & RMSE &  \textbf{0.840} &  0.854 & 0.856 & 0.846 \\
\midrule
Movie100k-binary & ACC & 0.717 & \textbf{0.722} & 0.692 & --\\
& AUC & 0.788 & \textbf{0.791} &\\
& MAP & 0.811 & \textbf{0.813}\\
Movie1M-binary & ACC & 0.739 & \textbf{0.746} & 0.732 & -- \\
& AUC & 0.809 & \textbf{0.818}\\
& MAP & 0.840 & \textbf{0.851}\\
Duolingo & ACC & \textbf{0.848} & 0.846 & 0.842 & --\\
& AUC & \textbf{0.814} & 0.809\\
& MAP & \textbf{0.948} & 0.946\\ \bottomrule
\vspace{-5pt}
\end{tabular}